\documentclass[10pt,twocolumn,letterpaper]{article}

\usepackage{cvpr}
\usepackage{times}
\usepackage{epsfig}
\usepackage{graphicx}
\usepackage{amsmath}
\usepackage{amssymb}

\usepackage{soul}
\usepackage{verbatim}
\usepackage{subfigure}
\usepackage{calc}
\usepackage{amstext}
\usepackage{stackengine}
\newcommand\IncG[2][]{\addstackgap{%
\raisebox{-.5\height}{\includegraphics[#1]{#2}}}}
\usepackage{array}
\usepackage{amsmath,amssymb} 
\usepackage{xcolor}
\usepackage{paralist}
\usepackage{multirow}
\usepackage{algorithm}  
\usepackage{algpseudocode}  
\usepackage{amsmath}  
\renewcommand{\algorithmicrequire}{\textbf{Input:}}

\usepackage[pagebackref=true,breaklinks=true,letterpaper=true,colorlinks,bookmarks=false]{hyperref}

\cvprfinalcopy 

\begin{document}
%
\title{Analysis of Deep Networks for Monocular Depth Estimation\\
Through Adversarial Attacks with Proposal of a Defense Method}

\author{ Junjie Hu$^{1,2}$ \hspace{1cm}  Takayuki Okatani$^{1,2}$\\
$^1$ Graduate School of Information Sciences, Tohoku University, Japan \\
$^2$ Center for Advanced Intelligence Project, RIKEN, Japan\\
{\tt\small \{junjie.hu, okatani\}@vision.is.tohoku.ac.jp}
}

\maketitle

\begin{abstract}{\color{black}
In this paper, we consider adversarial attacks against a system of monocular depth estimation (MDE) based on convolutional neural networks (CNNs). The motivation is two-fold. One is to study the security of MDE systems, which has not been actively considered in the community. The other is to improve our understanding of the computational mechanism of CNNs performing MDE. Toward this end, we apply the method recently proposed for visualization of MDE to defending attacks. It trains another CNN to predict a saliency map from an input image, such that the CNN for MDE continues to accurately estimate the depth map from the image with its non-salient part masked out. We report the following findings. First, unsurprisingly, attacks by IFGSM (or equivalently PGD) succeed in making the CNNs yield inaccurate depth estimates. Second, the attacks can be defended by masking out non-salient pixels, indicating that the attacks function by perturbing mostly non-salient pixels. However, the prediction of saliency maps is itself vulnerable to the attacks, even though it is not the direct target of the attacks. We show that the attacks can be defended by using a saliency map predicted by a CNN trained to be robust to the attacks. These results provide an effective defense method as well as a clue to understanding the computational mechanism of CNNs for MDE.
}
\end{abstract}

\section{Introduction}

Monocular depth estimation, i.e., estimating the depth of a three-dimensional scene from its single image, has a long history of research in the fields of computer vision and visual psychophysics. The recent employment of convolutional neural networks (CNNs) has gained significant improvement in the estimation accuracy \cite{Eigen2014depth,Li2015DepthAS,laina2016deeper,Kendall2017WhatUD,fu2018deep}. In this paper, we consider adversarial attacks to the CNNs performing this task and particularly defense methods against them. 

\begin{figure}[!t]
\centering  
\begin{tabular} 
{p{0.095\textwidth}<{\centering}p{0.095\textwidth}<{\centering}p{0.095\textwidth}<{\centering}p{0.095\textwidth}<{\centering}} 
\IncG[ width=0.7in]{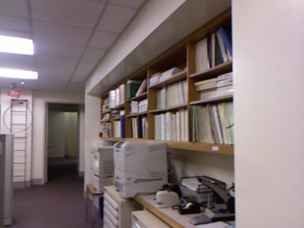}
&\IncG[ width=0.7in]{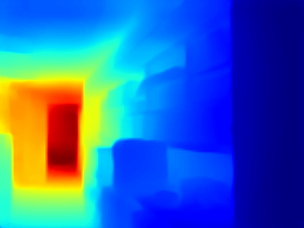}
&\IncG[ width=0.7in]{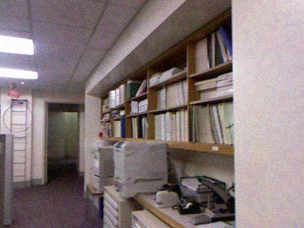}
&\IncG[ width=0.7in]{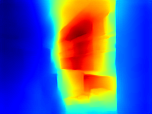}
\\
\IncG[ width=0.7in]{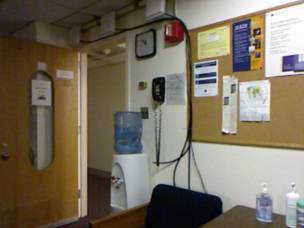}
&\IncG[ width=0.7in]{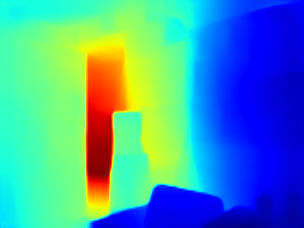}
&\IncG[ width=0.7in]{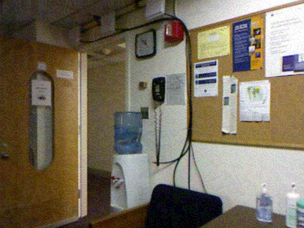}
&\IncG[ width=0.7in]{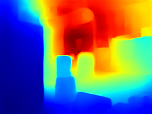}
\\
\IncG[ width=0.7in]{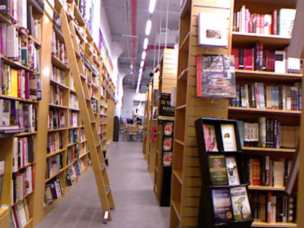}
&\IncG[ width=0.7in]{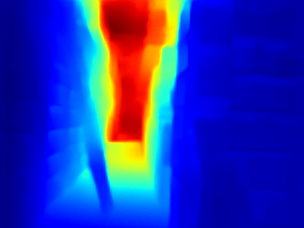}
&\IncG[ width=0.7in]{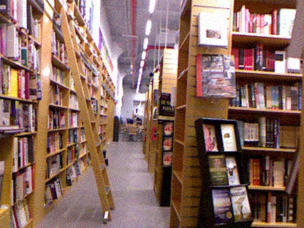}
&\IncG[ width=0.7in]{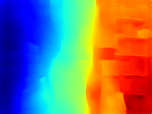}
\\
\footnotesize(a)&\footnotesize(b)&\footnotesize(c)&\footnotesize(d)
\end{tabular}
\vspace*{2mm}
\caption{Vulnerability of CNNs to adversarial examples on the monocular depth estimation task. (a) Input. (b) Estimated depth from (a). (c) Adversarial input created by IFGSM with  $\epsilon=0.1$ and 10 iterations; see Eq.(\ref{eq_ifgsm}) for details. (d) Estimate from (c). }
\label{fig_depth_attack}
\end{figure}

The motivation of this study is two-fold. First, we study the security of CNNs designed and trained for this task, which are beginning to be deployed in real-world applications.
Unsurprisingly, they are vulnerable to adversarial attacks, as shown in Fig.~\ref{fig_depth_attack}, although this has not been reported in the literature to the authors' knowledge. 

Second, we attempt to understand, through analyses of adversarial examples, in what computational mechanism these CNNs infer depth from a single image. Recently, a few studies have been published on this question, which indicates that these CNNs utilize similar cues to those that are believed to be used by the human vision \cite{Dijk_2019_ICCV,Hu2019VisualizationOC}. However, we are still on the way to a full understanding of the mechanism. If there are adversarial examples that lead the CNNs to malfunction, they should invalidate (some of) the cues that the CNNs utilize. By analyzing these,
we wish to gain a deeper understanding of the inference mechanism.  

Motivated by these two goals, we consider a defense method against adversarial attacks that is based on the recent study of Hu et al. \cite{Hu2019VisualizationOC} on the visualization of CNNs for the monocular depth estimation task. They propose a method for obtaining a saliency map, i.e.,  a set of a small number of pixels from which a CNN can estimate depth accurately. They show that such a saliency map can be predicted from the input image by another CNN trained for the purpose. They then show that CNNs can estimate a depth map fairly accurately from only a sparse subset of image pixels specified by the predicted saliency map. 

\begin{figure}[!t]
\centering  
\includegraphics[width=0.95\columnwidth]{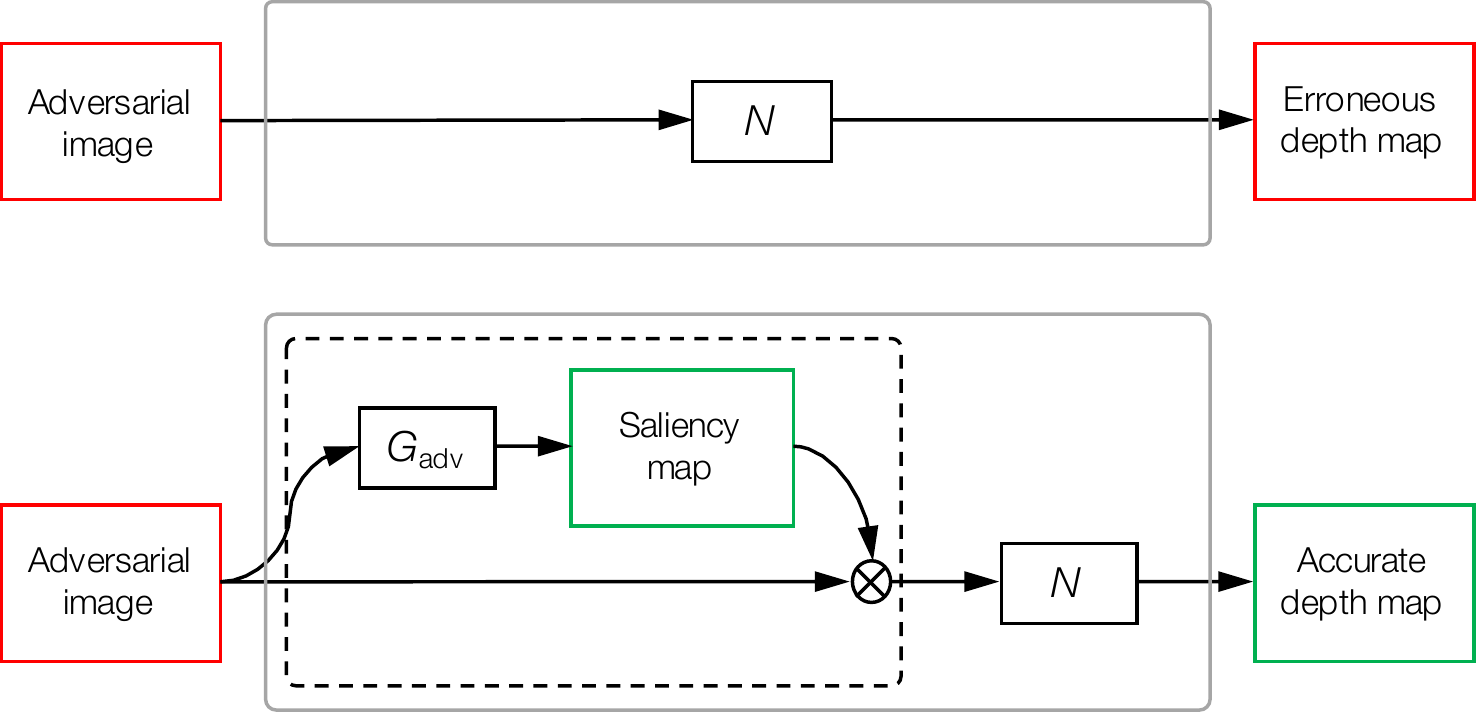}
\caption{Illustration of the proposed approach. Upper row: An adversarial image forces a depth estimator $N$ yield erroneous output. Lower row: The same adversarial image is detoxified by masking it with a saliency map, which is estimated by a robust estimator $G_{\mathrm{adv}}$. }
\label{fig_defense_method}
\end{figure}

We utilize their framework for visualization to construct a defense method, as illustrated in Fig.~\ref{fig_defense_method}. Given an adversarial input that will force the CNN $N$ to yield an erroneous depth map, a saliency map is first predicted by an auxiliary CNN $G$, which is multiplied with the (adversarial) input in a pixel-wise fashion; the resulting image is then fed to $N$, thereby aiming at obtaining an accurate depth map. 

This idea is based on the following considerations. Observing the saliency maps predicted by the method of Hu et al. \cite{Hu2019VisualizationOC}, we can say that the CNNs use the portions of images that are similar to those used by the human vision. Thus, it is natural to hypothesize that the CNNs use similar cues and perform similar computations based on them to the human vision. If so, why do the adversarial attacks still succeed? (Why is it possible to affect CNNs by adding slight perturbation to inputs that do not affect the human vision?) To reconcile these two, considering that the CNN receives all the pixels at their input, we conjecture that {\em the attacks are made possible not by perturbing salient pixels containing important depth cues but by mostly perturbing non-salient pixels.}

A recent study has made an intriguing argument about adversarial attacks on classification tasks \cite{Ilyas2019AdversarialEA}. There are a number of features in images that could be effective for classification {\em within} a given dataset. Dividing them into robust and non-robust features, the authors argue that the human vision uses only the former, whereas CNNs may use them both, and that adversarial attacks affect only the non-robust features, explaining why the attacks succeed without affecting recognition by the human vision. Our approach may be restated from their perspective that we define robust and non-robust `features' for monocular depth estimation by spatially dividing an input image based on the pixel saliency. 

\begin{figure*}[t]
\centering
\includegraphics[width=0.9\textwidth]{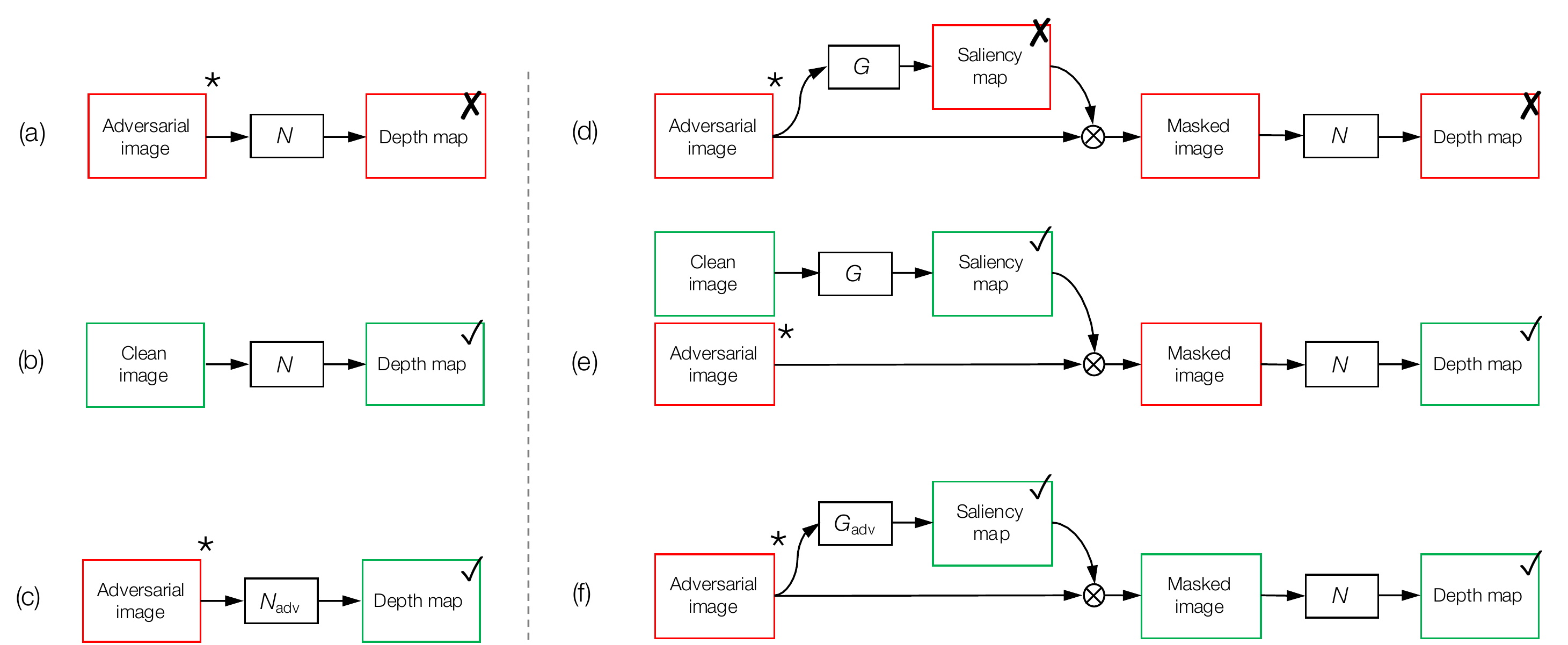}
\smallskip
\caption{Configurations of depth estimation network $N$ and saliency estimation network $G$ considered in this paper. (a) $N(x^*)$: $N$ is vulnerable to an adversarial input $x^*$. (b) $N(x)$: $N$ yields an accurate estimate from a clean image $x$. (c) $N_{\mathrm{adv}}(x^*)$: An estimator $N_{\mathrm{adv}}$ robust to $x^*$. (d) $N(x^*\otimes G(x^*))$: $N$ combined with $G$ remains to be vulnerable. (e) $N(x^*\otimes G(x))$: An accurate estimate is obtained when the clean image is input to $G$. (f) $N(x^*\otimes G_{\mathrm{adv}}(x^*))$: A robust estimator $G_{\mathrm{adv}}$ trained by using adversarial inputs can be used to defend the attack. All the adversarial images (with *) are identical; it is tailored for $N$ in setting (a).}
\label{fig:config}
\end{figure*}

In the rest of the paper, we will show the followings:
\begin{itemize}
    \item Adversarial attacks by IFGSM for a depth estimation CNN can be defended by masking out non-salient pixels in the inputs; see Fig.~\ref{fig:config}(e). This indicates that the attack functions mostly by perturbing non-salient pixels. 
    \item Attacks tailored for a depth prediction CNN also lead the auxiliary CNN (for saliency map prediction) to malfunction; see Fig.~\ref{fig:config}(d). 
    Thus, the attacks cannot be defended using the saliency maps predicted by them. 
    \item The attacks can be defended by immunizing the auxiliary CNN against the attacks, which is made possible by employing adversarial training for it;
    it is not necessary to make any change on the depth CNN. This will be a valid defense method; see Fig.~\ref{fig:config}(f).
\end{itemize}

\section{Backgrounds}

\subsection{Adversarial Attack and Defense}

In this paper, we mainly consider adversarial attacks by IFGSM \cite{Kurakin2017AdversarialEI}, the iterative version of the fast gradient sign method (FGSM) \cite{Goodfellow2015ExplainingAH}, although there are a number of attack methods; we refer the reader to a survey \cite{Akhtar2018ThreatOA}. This is because IFGSM, which is also known as projected gradient descent (PGD), is considered to be the strongest attack utilizing the local first order information about the network \cite{Madry2017TowardsDL}, and thus is primarily considered in the previous studies on defense methods. 

IFGSM is briefly summarized as follows. Suppose a network that has been trained for predicting $y$ from an input $x$ by minimizing a loss $\ell(x,y)$. Assuming we have access to the gradient of the loss, IFGSM iteratively updates inputs starting from a given input $x(\rightarrow x_0^*)$ according to 
\begin{equation}
\label{eq_ifgsm}
    \quad x^*_{t+1} = \text{Clip}_{\epsilon,x} \left\{ x^*_{t} +  \alpha
    \cdot \mathrm{sign}(\triangledown \ell(x_t{^*},y))\right\}.
\end{equation}
where $\epsilon$ denotes $l_\infty$ bound of adversarial perturbation and $\alpha$ is the step size.
As shown in Fig.~\ref{fig_depth_attack}, this method is effective for monocular depth estimation. For the purpose of clarification, we will use the following notation to indicate that an adversarial input $x^*$ is generated from a clean input $x$ targeting at a network $N$:
\begin{equation}    
    x^* = Adv(x; N),
\end{equation}
as shown in Fig.~\ref{fig:config}(a). 

There are many other studies on defense methods. They are roughly categorized into two classes. The first is to improve the robustness of the target network against adversarial images in some ways. The method of {\em adversarial training} \cite{Tramr2018EnsembleAT,Na2018CascadeAM} creates adversarial images and uses them for training along with normal images.
There are a number of methods that incorporate some regularizers to training, such as 
the employment of network distillation \cite{Hinton2015DistillingTK},  in which the prediction of another network is used as a soft label, the deep contrastive network, which employs a layer-wise contrastive penalty \cite{Gu2014TowardsDN}, and non-standard training schemes, such as saturating networks \cite{Nayebi2017BiologicallyIP}, parseval networks \cite{Ciss2017ParsevalNI} and adversarial-trained bayesian neural networks \cite{Liu2019AdvBNNIA}.

The other class of methods aim to remove perturbations from adversarial images before they are fed into the target network. It is proposed to detect adversarial images and then use an auto-encoder
to project them onto a learned manifold of clean images \cite{Meng2017MagNetAT}. A similar method is employed in Defense-GAN \cite{Samangouei2018DefenseGANPC}. 
It is also proposed to remove the perturbation from adversarial images by learning a denoiser network \cite{Liao2018DefenseAA}. 
The proposed method differs from any of the existing methods. It may be considered to lie in an intersection of the above two classes. 


\subsection{Visualizing CNNs on Depth Estimation}
\label{Hu_visualization}

It is a popular approach to visualize a saliency map,  aiming at understanding inference conducted by a CNN for an input \cite{Smilkov2017SmoothGradRN,Mahendran2016SalientDN,Sundararajan2017AxiomaticAF}. In the study of Hu et al. \cite{Hu2019VisualizationOC}, this is attempted for the monocular depth estimation task. Their method finds a set of a small number of pixels from which a CNN can estimate depth accurately. Specifically, 
they consider a mask $m$, or equivalently a saliency map, which is used to mask out the irrelevant pixels from $x$ by element-wise multiplication $x\otimes m$. Then, the goal is to find $m$ such that the original inference will be maintained as much as possible, that is, $N(x)\sim N(x\otimes m)$, and also $m$ is as sparse as possible, which reduces to the following optimization: 
\begin{equation}
    \min\limits_m \;
    \ell_{\rm dif} (N(x),N(x \otimes m)) + \lambda 
    \frac{1}{n} \lVert m \rVert_1,
    \label{eqn_M}
\end{equation}
where $\ell_{\rm dif}$ is a difference measure; 
$\lambda$ is a control parameter for the sparseness of $m$; $n$ is the number of pixels;  and $\lVert m\rVert_1$ is the $\ell_1$ norm (of a vectorized version) of $m$. To avoid artifacts generated by direct optimization of (\ref{eqn_M}), they propose to incorporate an auxiliary network $G$ and train it to predict $m$ from $x$ according to
\begin{equation}
    \min\limits_G \; \ell_{\rm dif} (y,N(x \otimes G(x)))+\lambda \frac{1}{n}
    \lVert G(x) \rVert_1.
\label{eqn_G}
\end{equation}
They show through experiments that the CNN $N$ continues to yield accurate depth from only a small number of pixels of $x$ that are delineated by the saliency map predicted from $x$ by $G$.

\color{black}

\section{Method}

\subsection{Saliency Map and Adversarial Perturbation}

We conducted a preliminary experiment to understand how the method of Sec.~\ref{Hu_visualization}, which was originally developed for visualization, reacts to adversarial attacks. In the experiment, we used the same network $N$ and $G$ as well as the dataset as in Sec.~\ref{white_box_experiments}. The results are shown in Table~\ref{attack_validation}. Here, adversarial inputs $x^*$'s are generated targeting at $N$, i.e., $x^*=Adv(x,N)$. The fist row, `$N(x^*)$', shows that $N$ is indeed vulnerable to this attack; error (RMSE) increases from 0.555 for $\epsilon=0$ (no attack) to 1.055 and 1.139 for $\epsilon=0.05$ and $0.1$, respectively. 

The second row, `$N(x^*\otimes G(x^*))$', shows the cases where the perturbed  $x^*$ is masked using the saliency map (predicted by $G$ from the same input $x^*$) and inputted to $N$. The results indicate that this configuration is similarly affected by the attack, although its effect is somewhat mitigated by the use of the mask. Note that the accuracy of this configuration $N(x^*\otimes G(x^*))$ is already a bit lower for clean images than $N(x^*)$, i.e., $0.683$ vs $0.555$. What comes as a surprise is that as shown in the third row, `$N(x^*\otimes G(x))$', when the mask $G(x)$ computed from a clean $x$ is applied to $x^*$, this immunizes the whole system to the attack. The increase of error is small, i.e., from $0.683$ to $0.696$ and $0.712$ for $\epsilon=0.05$ and $0.1$, respectively. 

These indicate that
\begin{itemize} \em
    \item FGSM generates an adversarial input $x^*=Adv(x,N)$ by perturbing mainly the non-salient pixels of $x$. 
    \item The adversarial inputs generated targeting at $N$ force the network $G$ to yield an inaccurate saliency map. 
\end{itemize}
Therefore, if we can make $G$ predict an accurate saliency map from $x^*$, or equivalently, $G(x^*)\sim G(x)$, then we will be able to defend the attacks. 

It should be noted here that if the attacker knows the configuration $C(x) \equiv N(x\otimes G(x))$ and can generate $x^{**}=Adv(x, C)$, this defense strategy may not work. In fact, our experiments showed that the error for $N(x^{**}\otimes G(x^{**})$ is 1.052 and 1.108 for $\epsilon=0.05$ and $0.1$, respectively. Thus, in what follows, we will consider the scenario where the attacker assumes that $N$ alone is used to predict depth.

\begin{table}[!t] 
\begin{center}
\caption{Results of FGSM attack on several configurations of the depth estimation network $N$  of \cite{Hu2019VisualizationOC}.  The numbers are RMSE values over the test set of NYU-v2; $x^*$ indicates the adversarial input given by $x^*=Adv(x; N)$; $G$ is the saliency estimation network. $N$ and $G$ are trained on clean images. $\epsilon=0$ means no attack. }
\label{attack_validation}\small
\begin{tabular}
{|p{.12\textwidth}<{\centering}|p{.08\textwidth}<{\centering}|p{.08\textwidth}<{\centering}|p{.08\textwidth}<{\centering}|}
\hline
 & $\epsilon=0$ & $\epsilon=0.05$ &$\epsilon=0.1$\\
\hline
 $N(x^*)$ & 0.555 &1.055  &1.139\\
 $N(x^*\otimes G(x^*))$ & 0.683 & 0.813&0.943\\
$N(x^*\otimes G(x))$& 0.683 & 0.696 &0.712\\
\hline
\end{tabular}
\end{center}
\end{table}
\setlength{\tabcolsep}{1.4pt}
\subsection{Saliency Prediction Robust to Attacks}

Thus, we wish to obtain the saliency prediction network $G$ that is robust to attacks. Then, our objective is two-fold. One is that, given an input image of a scene, we wish to identify as few pixels in the image as possible from which the depth can be predicted as accurately as possible. This leads to prediction of saliency map with a sparseness constraint, which is proposed in \cite{Hu2019VisualizationOC}. The other is that we want to make this prediction robust to possible adversarial attacks. 

To met the first requirement, we follow the method of $\cite{Hu2019VisualizationOC}$ for training with a slight modification. It is the use of ground truth depth $\bar{y}$ as the desired target; in the original method, the output $y=N(x)$ obtained from mask-free input is used as the target. We found in our experiments that this yields better performance. 

To satisfy the second requirement, we employ adversarial training; that is, we use not only clean images $x$'s but adversarially perturbed images $x^*$'s for training $G$. Note that we do not touch the depth prediction network $N$ in this process. 

Then, the training of $G$ is expressed as the following optimization:
\begin{equation}
    G_{\mathrm{adv}} = \mathop{\mathrm{argmin}}_G \; \ell_{\rm dif} (\bar{y},N(x' \otimes G(x')))+\lambda \frac{1}{n}
    \lVert G(x') \rVert_1,
\label{eqn_G2}
\end{equation}
where $x'$ is either an input clean image or an adversarial image generated from a clean image. 

\begin{algorithm}[t]  
  \caption{Algorithm for training the saliency prediction network $G$. The batch size is set to 1 for simplicity.}  
  \label{alg_leaning_G}  
  \begin{algorithmic}[1]  
    \Require  
      $N$: a target, fully-trained network for depth estimation;  
      $\chi$: a training set, \ie, pairs of an RGB image of a scene and its depth map;
      $\epsilon$: $\ell_\infty$ bound for IFGSM;
      $K$: training epochs; and
      $J$: iterations per epoch.
     \renewcommand{\algorithmicrequire}{\textbf{Hyperparameters:}}
    \Ensure  
      $G_{\mathrm{adv}}$: a network for predicting a saliency map.  
    \For{$k$ = 1 to $K$}
        \For{$j$ = 1 to $J$}
            \State Select an RGBD pair $\{x,\bar{y}\}$ from $\chi$
            \State $p=\mathrm{Uniform}(0,1)$ 
            \If{$p>0.5$}
                \State $\epsilon=\mathrm{Uniform}(0.01, 0.3)$
                \State $T=\lfloor \mathrm{Uniform}(1, 10) \rfloor$
                \State $x^*_0 = x$
                \State $t=0$
                \For{$t=1$ to $T$}
                \State $x^*_{t+1} = \mathrm{IFGSM}(x^*_{t},\epsilon)$
                \EndFor
            \State $x'=x^*_t$
            \Else
                \State $x'=x$
            \EndIf
            \State 
            $L = \ell_{\rm dif} (\bar{y},N(x' \otimes G(x')))+\lambda \frac{1}{n} \lVert G(x') \rVert_1$
             \State Backpropagate $L$
            \State Update $G$
        \EndFor
    \EndFor 
    \State $G_{\mathrm{adv}}\leftarrow G$
  \end{algorithmic}  
\end{algorithm}

The details of the procedure of training are given in Algorithm~\ref{alg_leaning_G}. The adversarial examples $x^*$'s are generated by IFGSM so that prediction error measured by $\ell_1$ norm of depth map will be maximized. The $\ell_\infty$ bound $\epsilon$ and the number of iterations for IFGSM are randomly chosen from uniform distributions in the range of 
$(0.01, 0.3)$ and
$(1, 10)$, respectively. 
 Following most of the previous studies, the step size $\alpha$ in Eq.(\ref{eq_ifgsm}) is set to 1. Thus, if the number of iterations is one, it reduces to FGSM.
For $\ell_{\mathrm{dif}}$, we use the loss function proposed in \cite{hu2019revisiting}, which calculates the errors of depth, gradient, and normals, as:
\begin{equation}
    \ell_{\rm dif} = l_{\rm depth} + l_{\rm grad} + l_{\rm normal},
\end{equation}
where $l_{\rm depth}=\frac{1}{n}\sum_{i=1}^n F(e_i)$ with $F(e_i) = \ln(e_i+0.5)$, $e_i = \|\bar{y}_i - y_i\|_1$, and  $\bar{y}_i$ and $y_i$ are true and estimated depths; 
$l_{\rm grad}=\frac{1}{n}\sum_{i=1}^n (F(\nabla_{x}(e_i))+F(\nabla_{y}(e_i)))$; and $ l_{\rm normal} = \frac{1}{n}\sum_{i=1}^n \left(1-\cos\theta_i\right)$ where $\theta_i$ is the angle between the surface normals computed from the true and estimated depth map.

\subsection{Defense by Masking Least Salient Pixels}

We expect that the above procedure will yield $G_{\mathrm{adv}}$ that can robustly predict a correct saliency map even from an adversarial input $x^*=Adv(x, N)$, i.e., $G_{\mathrm{adv}}(x^*)\sim G(x)$. Recall here that we do not touch the depth estimation network $N$ during the training of $G_{\mathrm{adv}}$. 
This is different from a standard defense strategy that tries to make $N$ itself robust to attacks, such as employment of adversarial training on $N$. This could be an advantage in some practical applications. In this sense, our method is more similar to the `denoising' approach to cope with attacks, such as projection of adversarial images to the manifold of clean images \cite{Liao2018DefenseAA}. However, it also differs from them in that it identifies and masks irrelevant pixels for depth estimation,  thereby minimizing the impact of attacks. 


\section{Experiments}

\subsection{Experimental Setting}

\paragraph{Dataset}
We use the NYU-v2 dataset \cite{Silberman2012IndoorSA} for all the experiments. The dataset consists of a variety of indoor scenes and is  used in most of the previous studies. We use the standard  procedure  for preprocessing \cite{Eigen2014depth,laina2016deeper,ma2017sparse}. To be specific, the official splits of 464 scenes are used, i.e., 249 scenes for training and 215 scenes for testing. The official toolbox is used to extract RGB images and depth maps from the raw data, and then fill in missing pixels in the depth maps to generate ground truths. This results in approximately 50K unique pairs of an image and a depth map of 640$\times$480 pixel size. The images are then resized down to 320$\times$240 pixels using bilinear interpolation, and then crop their central parts of 304$\times$228 pixels, which are used as inputs to networks. The depth maps are resized to 
$152\times 114$ pixels. 
For testing, following the previous studies, we use the same small subset of 654 samples.

\paragraph{Networks} 
For $N$, we use the network built on ResNet-50 in \cite{hu2019revisiting}. It is trained on the data mentioned above. 
For $G_{\mathrm{adv}}$ (and $G$), we employ the same encoder-decoder network as \cite{Hu2019VisualizationOC}. It employs a dilated residual network (DRN) proposed in \cite{Yu2017} for the encoder part and three stacks of the up-projection blocks proposed in \cite{laina2016deeper} for the decoder part. It outputs a saliency map of the same size as the input image. 
We train $G_{\mathrm{adv}}$ according to Algorithm 1 for 60 epochs. The parameter $\lambda$ controlling the sparseness of the saliency map is set to 1. We use the Adam optimizer with a learning rate of 0.0001, $(\beta_{1},\beta_{2})=(0.9,0.999)$ and weight decay of 0.0001. 

\subsection{Defense to White-box Attacks}
\label{white_box_experiments}

\begin{figure}[t]
\centering
\subfigure {\includegraphics[width=60mm]{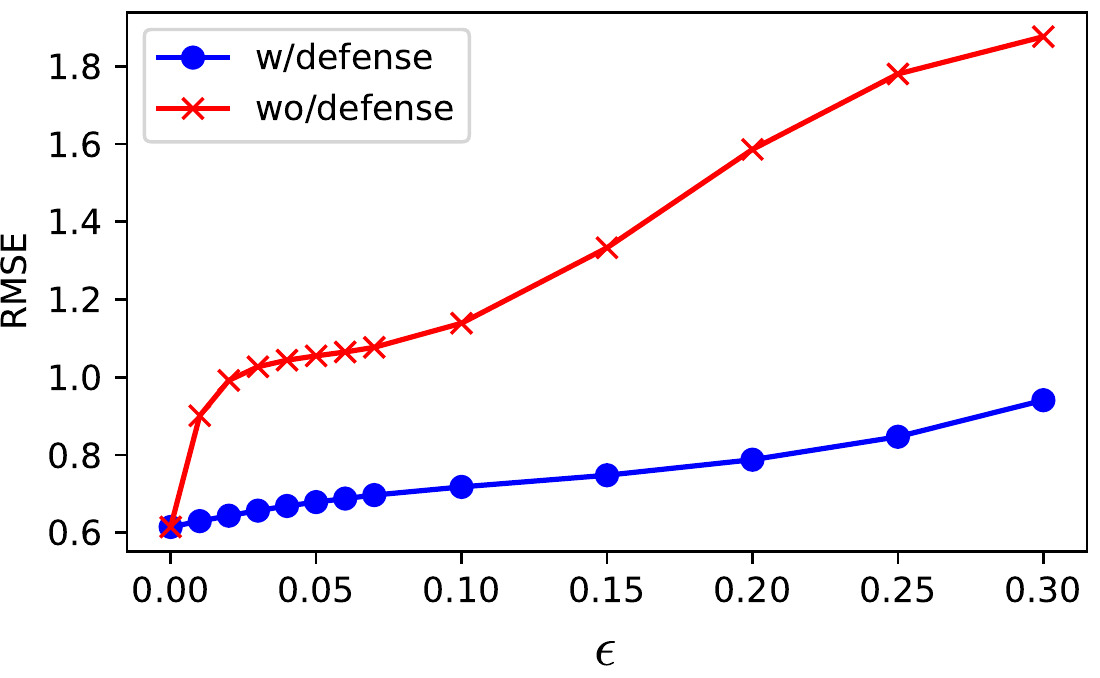}}
\caption{The RMSE of the depth maps estimated by $N(x^*)$ (`w/o defense') and $N(x^*\otimes G_{\mathrm{adv}}(x^*))$ (`w/defense') from adversarial inputs generated by FGSM with different perturbation strength $\epsilon$. } 
\label{fig_white_attack}
\end{figure}

\paragraph{FGSM attacks with different $\epsilon$'s}
We first evaluated the performance of the plain network $N(x^*)$ (Fig.~\ref{fig:config}(a)) and the proposed method $N(x^*\otimes G_{\mathrm{adv}}(x^*))$ (Fig.~\ref{fig:config}(f)) for FGSM attacks with different magnitude $\epsilon$ of perturbation. Each adversarial example $x^*$ is generated by FGSM with an $\epsilon$ to maximize the $\ell_1$ norm of the difference between the estimate $N(x^*)$ and its ground truth. We generated one $x^*$ for each of the NYU-v2 test images and calculate the average of depth estimation errors (measured by RMSE) by the two models. The results are shown in Fig.~\ref{fig_white_attack}. 
It is first confirmed from the plot that the attack is indeed effective for the plain network $N(x^*)$ without any defense; the error increases rapidly even with small $\epsilon$. It is also seen that our method $N( x^*\otimes G_{\mathrm{adv}}(x^*) )$ works fairly well; the error only increases slowly with increasing $\epsilon$. 

\begin{figure}[t]
\centering
\subfigure {\includegraphics[width=60mm]{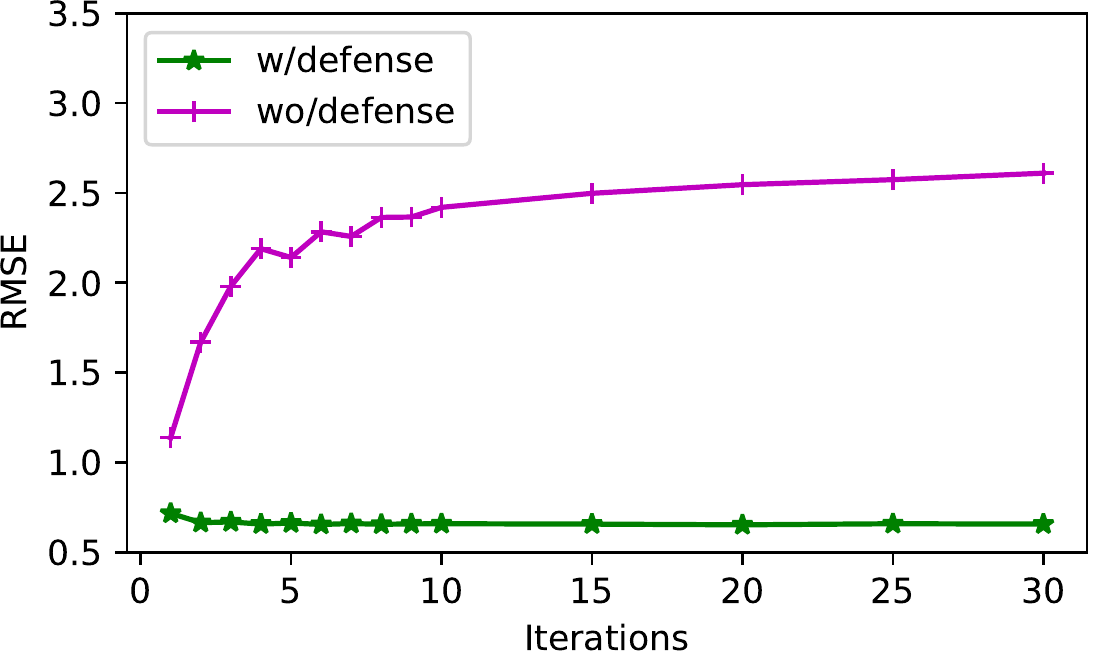}}
\caption{The RMSE of the depth maps estimated by the same two methods from adversarial inputs generated by IFGSM with different iterations.} 
\label{fig_rmse_iterations}
\end{figure}

\begin{table*}[t]
\begin{center}
\caption{Quantitative comparisons of four depth estimation methods. $N(x)$ is the plain network trained only on clean images. $N_{\mathrm{adv}}(x)$ is the same net but trained also on adversarial inputs. $G(x)$ is the saliency predictor trained only on clean images. $G_{\mathrm{adv}}(x)$ is the same net but trained on adversarial inputs by Algorithm \ref{alg_leaning_G}. The adversarial inputs $x^*$'s are generated targeting $N(x)$ by IFGSM with 10 iterations.} \small
\label{quan_res}
\smallskip
\begin{tabular}{|l|l|c|c|c|c|c|c|} 
\hline 
Attack&Prediction Method & RMSE $\downarrow$ & REL$\downarrow$ &$\log10$ $\downarrow$ &$\delta<1.25$ $\uparrow$ &$\delta<1.25^{2}$ $\uparrow$& $\delta<1.25^{3}$ $\uparrow$\\

\hline
No attack (clean)
& $N(x)$  &\textbf{0.555} &\textbf{0.126}&\textbf{0.054} &\textbf{0.843} &\textbf{0.968} &\textbf{0.991}\\
&$N_{\mathrm{adv}}(x)$  &0.682 &0.167 &0.071 &0.757 &0.935 &0.979\\
&
$N(x\otimes G(x))$
&0.683&0.154&0.068&0.773&0.939&0.982\\
&$N(x\otimes G_{\mathrm{adv}}(x))$ &{0.615} &{0.148} &{0.063} &{0.792} &{0.952} &{0.987}\\

\hline
IFGSM ($\epsilon=0.05$)
&$N(x^*)$  &1.465&0.419 &0.200&0.249&0.568&0.774\\
&$N_{\mathrm{adv}}(x^*)$   &0.666 &0.160 &0.067 &\textbf{0.774} &0.943 &0.982\\

&$N(x^*\otimes G(x))$ &0.692&\textbf{0.156} &0.069&0.768&0.937&0.981\\

&$N(x^*\otimes G_{\mathrm{adv}}(x^*))$&\textbf{0.644} &0.158 &\textbf{0.067} &0.771 &\textbf{0.945} &\textbf{0.984}\\

\hline
IFGSM ($\epsilon=0.1$)
&$N(x^*)$  &1.792&0.373 &0.273&0.161&0.373&0.571\\
&$N_{\mathrm{adv}}(x^*)$  &0.677 &0.159 &0.068 &0.769 &0.942 & 0.981\\
&$N(x^*\otimes G(x))$ &0.706&\textbf{0.158} &0.070&0.763&0.934&0.980\\
&$N(x^*\otimes G_{\mathrm{adv}}(x^*))$&\textbf{0.655} &0.160 &\textbf{0.067} &\textbf{0.770} &\textbf{0.942} &\textbf{0.983} \\
\hline

IFGSM ($\epsilon=0.15$)
&$N(x^*)$  &1.988& 0.516&0.325&0.109&0.263&0.442\\
&$N_{\mathrm{adv}}(x^*)$   &0.724 &0.167 &0.073 &0.741 &0.931 &0.976 \\
&$N(x^*\otimes G(x))$ &0.720&\textbf{0.159} &0.071&0.759&0.931&0.978\\
&$N(x^*\otimes G_{\mathrm{adv}}(x^*))$ &\textbf{0.677} &0.162 & \textbf{0.068} & \textbf{0.767} & \textbf{0.939} & \textbf{0.981}\\
\hline

IFGSM ($\epsilon=0.2$)
&$N(x^*)$  &2.107 &0.541 &0.360 &0.075 &0.201 &0.370 \\
&$N_{\mathrm{adv}}(x^*)$ &0.798 &0.180 &0.081 &0.701 &0.911 &0.970\\
&$N(x^*\otimes G(x))$ &0.743&\textbf{0.161} &0.074&0.751&0.927&0.977\\
&$N(x^*\otimes G_{\mathrm{adv}}(x^*))$ &\textbf{0.703} &0.165 & \textbf{0.071} &\textbf{0.754} &\textbf{0.933} & \textbf{0.979}\\
\hline
\end{tabular}
\end{center}
\end{table*}

\paragraph{IFGSM attacks}
We also evaluated their performance for IFGSM attacks. Figure \ref{fig_rmse_iterations} shows the averaged RMSE over all the inputs $x^*$'s generated by IFGSM with various iterations and a constant $\epsilon=0.1$. It is seen that the plain network $N(x^*)$ suffers more from $x^*$ generated with larger iterations, whereas the proposed method $N( x^*\otimes G_{\mathrm{adv}}(x^*) )$ is able to stably defend the attack. Although it is not shown here, it continues to perform well for larger iterations ($>$ 10).

We then compared the above two methods and additionally two other methods on IFGSM attacks with different perturbation magnitude. One additional method is 
the network $N$ directly trained by adversarial training \cite{Kurakin2017AdversarialEI}, denoted by $N_{\mathrm{adv}}$ (Fig.~\ref{fig:config}(c)). 
 Specifically, similarly to the training of $G_{\mathrm{adv}}$, it is trained using both clean and adversarial images, but its encoder part (Resnet-50) starts with a pretrained model on ImageNet, following \cite{hu2019revisiting}. The other is $N(x^*\otimes G(x))$ (Fig.~\ref{fig:config}(e)), i.e., inputting to the plain network $N$ the multiplication of $x^*$ and a saliency map predicted from a clean input $x$ by $G$. Here, $G$ is trained using only clean images, for which the sparseness parameter is set to $\lambda=5$ as recommended in \cite{Hu2019VisualizationOC}.

\begin{figure*}[t]
\centering  
\begin{tabular} 
{p{0.115\textwidth}<{\centering}p{0.115\textwidth}<{\centering}p{0.115\textwidth}<{\centering}p{0.115\textwidth}<{\centering}p{0.115\textwidth}<{\centering}p{0.115\textwidth}<{\centering}p{0.115\textwidth}<{\centering}p{0.115\textwidth}<{\centering}}
\IncG[ width=0.78in]{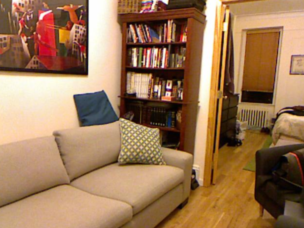}
&\IncG[ width=0.78in]{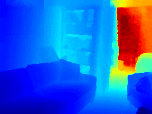}
&\IncG[ width=0.78in]{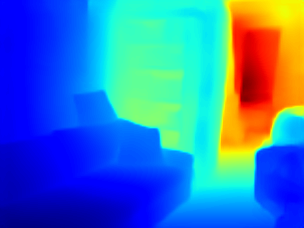}
&\IncG[ width=0.78in]{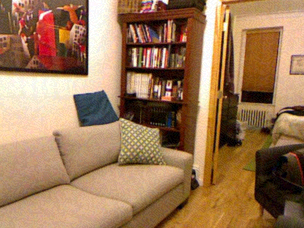}
&\IncG[ width=0.78in]{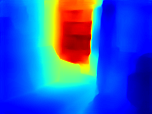}
&\IncG[ width=0.78in]{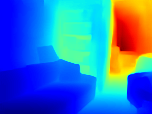}
&\IncG[ width=0.78in]{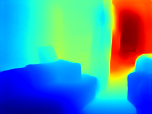}
&\IncG[ width=0.78in]{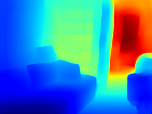}
\\
\IncG[ width=0.78in]{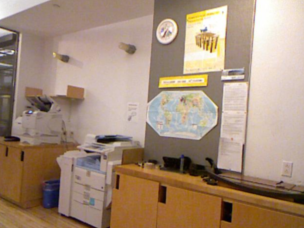}
&\IncG[ width=0.78in]{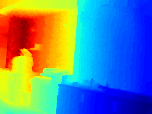}
&\IncG[ width=0.78in]{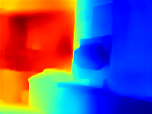}
&\IncG[ width=0.78in]{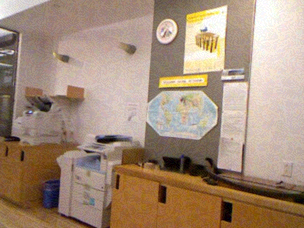}
&\IncG[ width=0.78in]{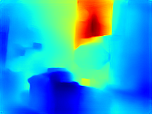}
&\IncG[ width=0.78in]{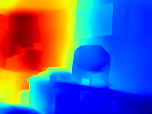}
&\IncG[ width=0.78in]{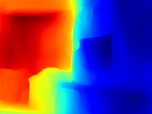}
&\IncG[ width=0.78in]{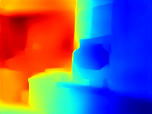}
\\
\IncG[ width=0.78in]{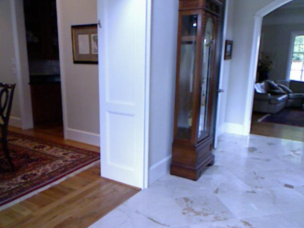}
&\IncG[ width=0.78in]{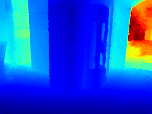}
&\IncG[ width=0.78in]{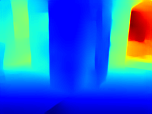}
&\IncG[ width=0.78in]{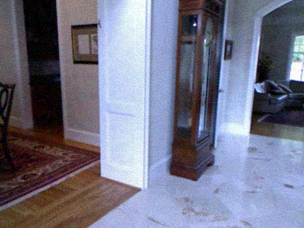}
&\IncG[ width=0.78in]{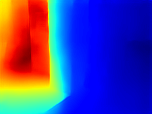}
&\IncG[ width=0.78in]{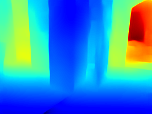}
&\IncG[ width=0.78in]{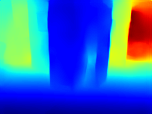}
&\IncG[ width=0.78in]{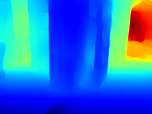}
\\
\IncG[ width=0.78in]{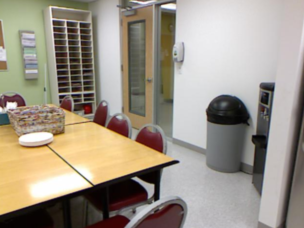}
&\IncG[ width=0.78in]{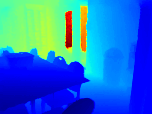}
&\IncG[ width=0.78in]{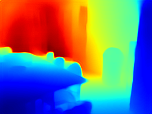}
&\IncG[ width=0.78in]{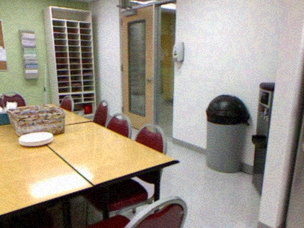}
&\IncG[ width=0.78in]{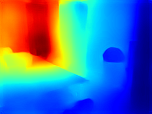}
&\IncG[ width=0.78in]{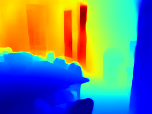}
&\IncG[ width=0.78in]{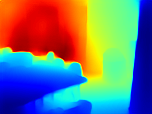}
&\IncG[ width=0.78in]{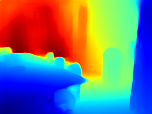}
\\
\IncG[ width=0.78in]{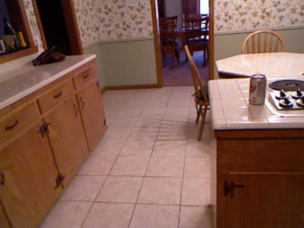}
&\IncG[ width=0.78in]{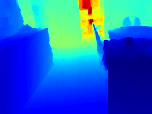}
&\IncG[ width=0.78in]{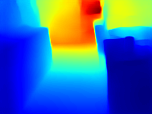}
&\IncG[ width=0.78in]{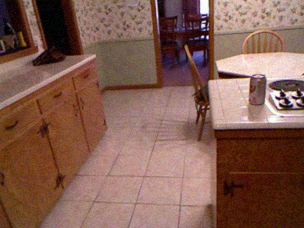}
&\IncG[ width=0.78in]{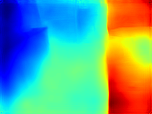}
&\IncG[ width=0.78in]{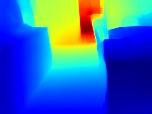}
&\IncG[ width=0.78in]{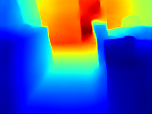}
&\IncG[ width=0.78in]{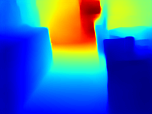}
\\
\IncG[ width=0.78in]{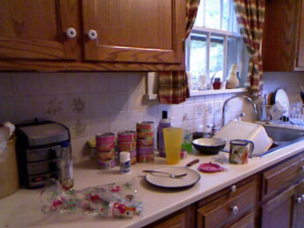}
&\IncG[ width=0.78in]{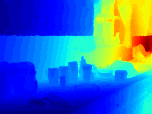}
&\IncG[ width=0.78in]{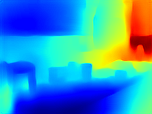}
&\IncG[ width=0.78in]{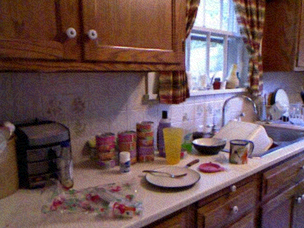}
&\IncG[ width=0.78in]{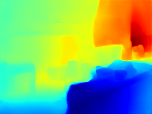}
&\IncG[ width=0.78in]{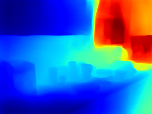}
&\IncG[ width=0.78in]{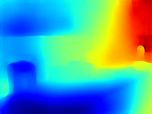}
&\IncG[ width=0.78in]{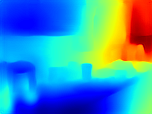}
\\
{\footnotesize (a) $x$}
& {\footnotesize (b) True depth}
& {\footnotesize (c) $N(x)$}
&{\footnotesize (d)  $x^*$}
&{\footnotesize (e)  $N(x^*)$} 
&{\footnotesize (f)  $N_{\mathrm{adv}}(x^*)$} 
&{\footnotesize (g)  $N(\!x^*\!\otimes\! G(x)\!)$} 
&{\footnotesize (h)  $N(\!x^*\!\otimes\! G_{\mathrm{adv}}(\!x^*\!)\!)$}
\end{tabular}
\vspace*{2mm}
\caption{Visual comparisons of depth maps estimated from adversarial inputs $x^*$'s generated by IFGSM with $\epsilon=0.1$ and 10 iterations; $x$'s are clean inputs.}
\label{fig_depth_nyu}
\end{figure*}

Table \ref{quan_res} shows the evaluation of the four methods
using multiple accuracy metrics. Adversarial examples were generated by IFGSM with four different $\epsilon$'s and 10 iterations. It is observed from the table that the three methods are all effective in defending the attacks for all the $\epsilon$'s. It is also seen that the proposed method, $N(x^*\otimes G_{\mathrm{adv}}(x^*))$, achieves the best performance among the three; it is better than $N_{\mathrm{adv}}(x^*)$ by a good margin and is equivalent or slightly better than $N(x^*\otimes G(x))$. Note that $N(x^*\otimes G(x))$ cannot be used in practice, as it needs a clean $x$. It is an ideal implementation of our conjecture that an adversarial input $x^*$ can be {\em detoxified} by masking $x^*$ with an accurate saliency map. The results show that this ideal implementation works well as a defense method, which well validates our conjecture. The results also show that the proposed method achieves comparable to or even better performance than it, which confirms that $G_{\mathrm{adv}}(x^*)$ can yield accurate saliency map even from $x^*$ and help defense to attacks. 
Figure~\ref{fig_depth_nyu} shows the depth maps predicted by these methods from adversarial inputs for different scenes. 


\paragraph{Effects of Losses Used for Adversarial Example Generation}
Adversarial inputs are generated so that they will maximize errors.  
There are a number of choices in the measure of the errors, or losses, such as $\ell_1$ norm (used in the experiments so far), $\ell_2$, etc. 
To investigate if the defending performance will be affected by the choice of this loss, we conducted an experiment. Specifically, we use the same $G_{\mathrm{adv}}$ trained assuming $\ell_1$ and test its performance against adversarial images generated using $\ell_2$, REL, $\log10$ and $\ell_{\rm dif}$, respectively. Here we use IFGSM with 5 iterations and $\epsilon=0.1$. Table~\ref{attack_losses} shows the results, indicating that all the losses are effective in making the plain network $N(x^*)$ yield erroneous depths, whereas the proposed method $N(x^*\otimes G_{\mathrm{adv}}(x^*))$ consistently performs well independently of the loss used for the adversarial example generation. 
\setlength{\tabcolsep}{3.2pt}
\begin{table}[t]
\begin{center}
\caption{Effects of the use of different losses for creation of adversarial inputs. $G_{\mathrm{adv}}$ is a single model trained assuming $\ell_1$. RMSE of depth estimation. }
\label{attack_losses}\small
\smallskip
\begin{tabular}
{|p{.08\textwidth}<{\centering}|p{.09\textwidth}<{\centering}|p{.15\textwidth}<{\centering}|}
\hline
Loss & $N(x^*)$ & $N(x^*\otimes G_{\mathrm{adv}}(x^*))$ \\
\hline
$\ell_1$ &1.778 &0.664 \\
$\ell_2$ &1.842 &0.666 \\
REL &1.694 &0.664 \\
$\log10$ &1.783 & 0.659 \\
$\ell_{\rm dif}$ &1.718&0.662\\
\hline
\end{tabular}
\end{center}
\end{table}
\setlength{\tabcolsep}{1.4pt}

\paragraph{Effects of Number of Layers of $G_{\mathrm{adv}}$}
We also investigate how the number of layers of the saliency estimation network $G_{\mathrm{adv}}$ affects defense performance. We tested $G_{\mathrm{adv}}$ having different numbers of layers for its encoder part: 22, 38, and 54 layers. Adversarial examples are generated by IFGSM with 5 iterations and $\epsilon=0.1$. Table~\ref{layer_depth} shows the results. 
It indicates that the more layers $G_{\mathrm{adv}}$ has, the better  performance we have, although the improvements are fairly small.

\setlength{\tabcolsep}{3.2pt}
\begin{table}[t]
\begin{center}
\caption{Effects of the number of layers of $G_{\mathrm{adv}}$ on adversarial defense.} \small
\label{layer_depth}
\smallskip
\begin{tabular}
{|p{.24\textwidth}<{\centering}|p{.12\textwidth}<{\centering}|}
\hline
Layers of the encoder in $G_{\mathrm{adv}}$ & RMSE \\
\hline
W/o $G_{\mathrm{adv}}$ (plain $N$)
&1.778\\
\hline
22&0.664\\
38&0.659\\
54&0.655\\
\hline
\end{tabular}
\end{center}
\end{table}
\setlength{\tabcolsep}{1.4pt}

\begin{figure*}[!th]
\label{saliency_map}
\centering  
\begin{tabular} 
{p{0.115\textwidth}<{\centering}p{0.115\textwidth}<{\centering}p{0.115\textwidth}<{\centering}p{0.115\textwidth}<{\centering}p{0.115\textwidth}<{\centering}p{0.115\textwidth}<{\centering}p{0.115\textwidth}<{\centering}p{0.115\textwidth}<{\centering}p{0.115\textwidth}<{\centering}}
\IncG[ width=0.78in]{./figs/adv_imgs_ep01_ite10/img65.png}
&\IncG[ width=0.78in]{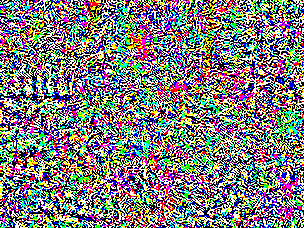}
&\IncG[ width=0.78in]{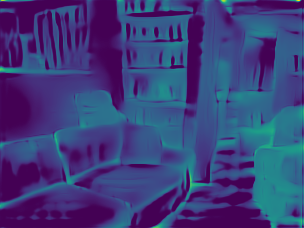}
&\IncG[ width=0.78in]{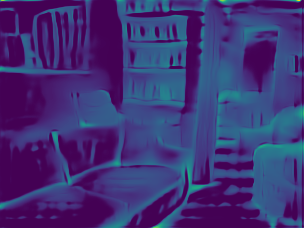}
&\IncG[ width=0.78in]{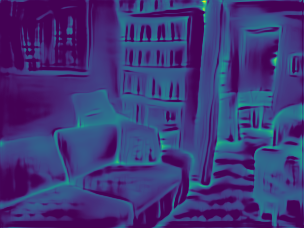}
&\IncG[ width=0.78in]{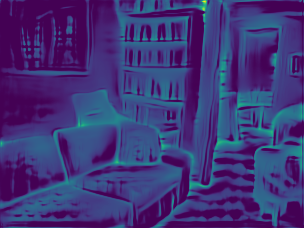}
&\IncG[ width=0.78in]{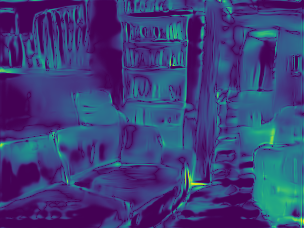}
&\IncG[ width=0.78in]{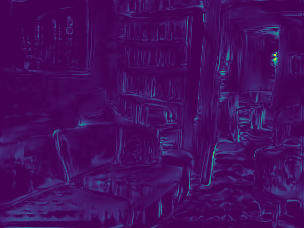}
\\
\IncG[ width=0.78in]{./figs/adv_imgs_ep01_ite10/img18.png}
&\IncG[ width=0.78in]{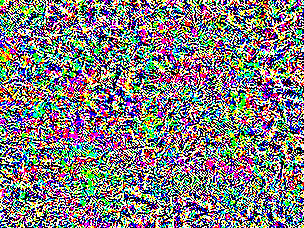}
&\IncG[ width=0.78in]{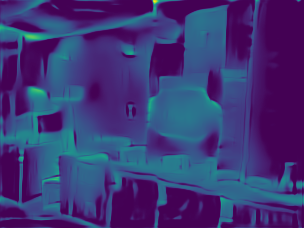}
&\IncG[ width=0.78in]{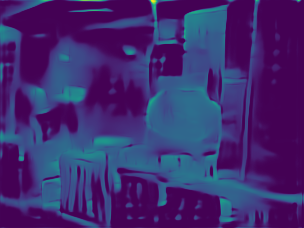}
&\IncG[ width=0.78in]{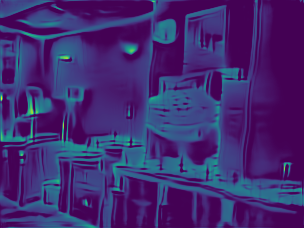}
&\IncG[ width=0.78in]{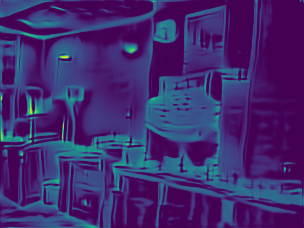}
&\IncG[ width=0.78in]{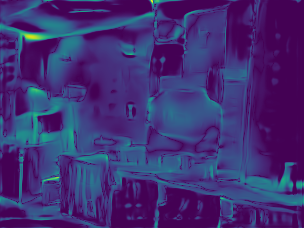}
&\IncG[ width=0.78in]{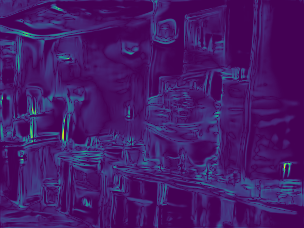}
\\
\IncG[ width=0.78in]{./figs/adv_imgs_ep01_ite10/img139.png}
&\IncG[ width=0.78in]{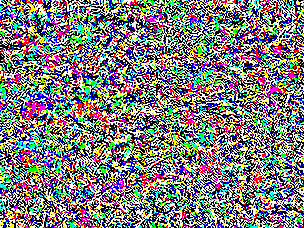}
&\IncG[ width=0.78in]{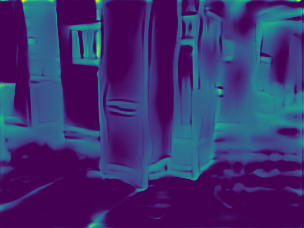}
&\IncG[ width=0.78in]{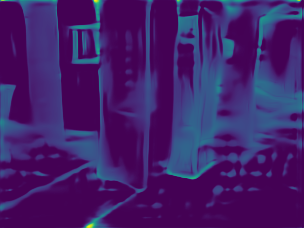}
&\IncG[ width=0.78in]{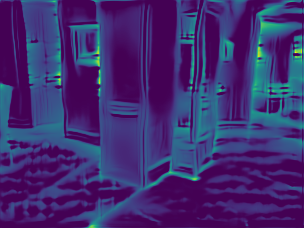}
&\IncG[ width=0.78in]{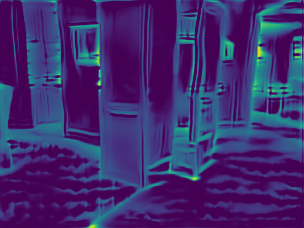}
&\IncG[ width=0.78in]{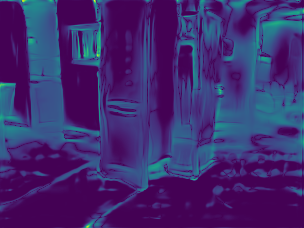}
&\IncG[ width=0.78in]{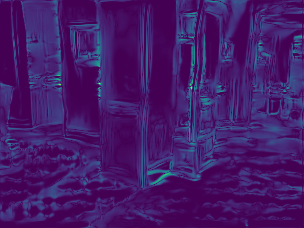}
\\
\IncG[ width=0.78in]{./figs/adv_imgs_ep01_ite10/img164.png}
&\IncG[ width=0.78in]{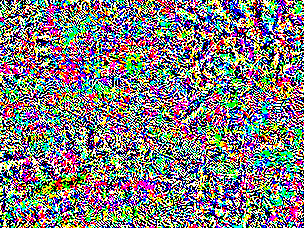}
&\IncG[ width=0.78in]{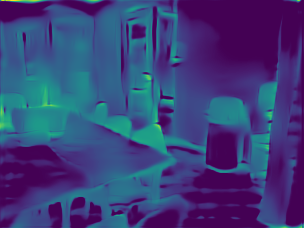}
&\IncG[ width=0.78in]{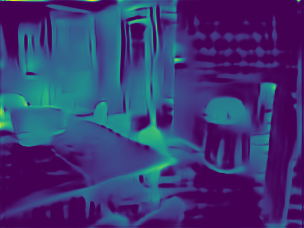}
&\IncG[ width=0.78in]{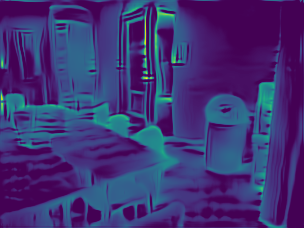}
&\IncG[ width=0.78in]{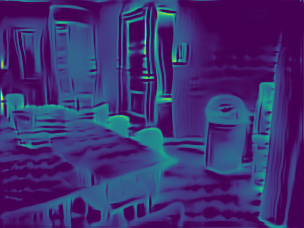}
&\IncG[ width=0.78in]{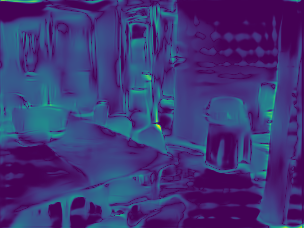}
&\IncG[ width=0.78in]{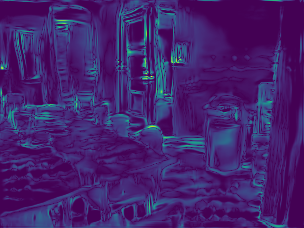}
\\
\IncG[ width=0.78in]{./figs/adv_imgs_ep01_ite10/img317.png}
&\IncG[ width=0.78in]{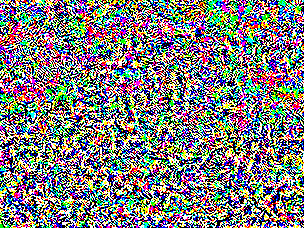}
&\IncG[ width=0.78in]{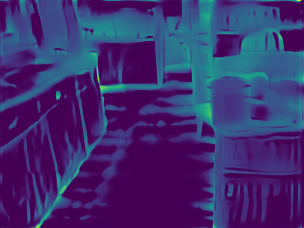}
&\IncG[ width=0.78in]{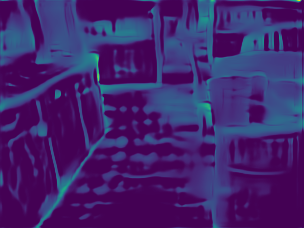}
&\IncG[ width=0.78in]{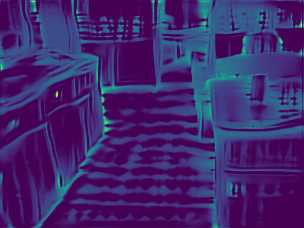}
&\IncG[ width=0.78in]{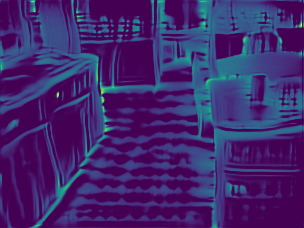}
&\IncG[ width=0.78in]{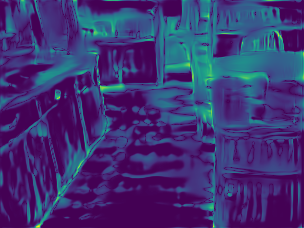}
&\IncG[ width=0.78in]{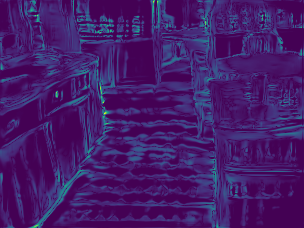}
\\
\IncG[ width=0.78in]{./figs/adv_imgs_ep01_ite10/img226.png}
&\IncG[ width=0.78in]{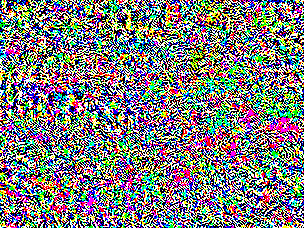}
&\IncG[ width=0.78in]{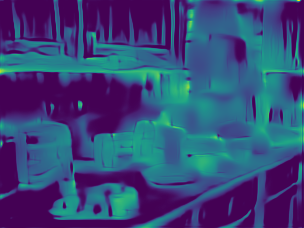}
&\IncG[ width=0.78in]{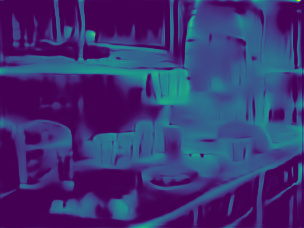}
&\IncG[ width=0.78in]{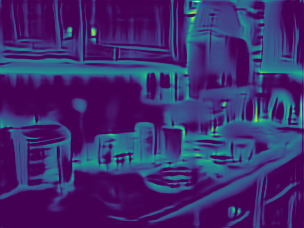}
&\IncG[ width=0.78in]{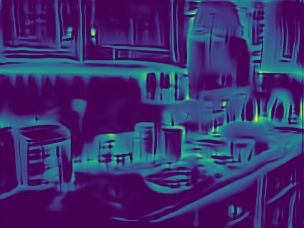}
&\IncG[ width=0.78in]{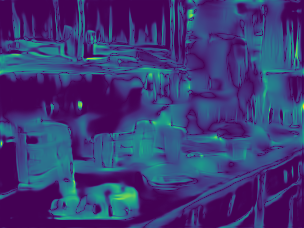}
&\IncG[ width=0.78in]{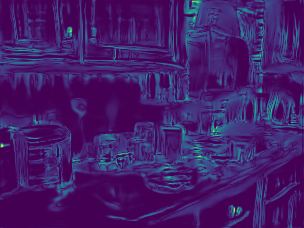}
\\
 {\footnotesize (a) $x^*$}
 &{\footnotesize (b) $x-x^*$} 
 & {\footnotesize (c) $G(x)$} 
 & {\footnotesize (d) $G(x^{*})$}  
 & {\footnotesize (e) $G_{\mathrm{adv}}(x)$}
 & {\footnotesize (f) $G_{\mathrm{adv}}(x^{*})$}
 &    {\footnotesize (g) $|G(x)-G(x^{*})|$}
 &{\footnotesize (h) $|G_{\mathrm{adv}}(x)-G_{\mathrm{adv}}(x^{*})|$} 
\end{tabular}
\caption{Visualization of clean and adversarial inputs and the saliency maps predicted from them. An identical color map created on a certain range is used for (c)-(f) and for (g) and (h), respectively. }
\label{fig_saliency_nyu}
\end{figure*}

\subsection{Qualitative Analysis}

Figure \ref{fig_saliency_nyu} shows adversarial examples and the saliency maps from them predicted by $G$ and $G_{\mathrm{adv}}$, along with their differences showing the effects of attacks. Here, the same images as Fig.~\ref{fig_depth_nyu} are chosen, for which adversarial inputs are generated by IFGSM with $\epsilon=0.1$ and 10 iterations. It is first observed that the perturbations (Fig.~\ref{fig_saliency_nyu}(b)) caused by the attack do not appear to be correlated with the saliency map and rather appear to be random.
Although this agrees with the previously reported observation (mostly on classification tasks), 
it may not appear to be consistent with our hypothesis that adversarial examples function by perturbing non-salient pixels. This may be because that in the attack by FGSM/IFGSM (Eq.(\ref{eq_ifgsm})), a constraint is imposed only on the magnitude of perturbation but not on the number of perturbed pixels. On the other hand, it can be confirmed from $|G(x)-G(x^*)|$ (Fig.~\ref{fig_saliency_nyu}(g)) that the attacks affect $G$ in its prediction, although $x^*$ is created for $N$ not for $G$. It is also seen from $|G_{\mathrm{adv}}(x)-G_{\mathrm{adv}}(x^*)|$ (Fig.~\ref{fig_saliency_nyu}(h)) that $G_{\mathrm{adv}}$ is indeed robust to the attacks. 


\section{Conclusion}

In this paper, we have analyzed how CNNs for monocular depth estimation react against adversarial attacks. 
For the IFGSM attack, we have validated the three items listed at the last of Sec.~1 through a number of experiments. They can be summarized as follows. IFGSM attacks can be defended by masking out non-salient pixels in the inputs. The non-salient pixels can be identified either by predicting saliency maps from clean images or by predicting saliency maps from adversarial inputs using an estimator trained to be robust to the attacks. This study has two contributions. One is the proposal of an effective defense method against the IFGSM attack, provided that the attacker has the knowledge of the depth estimation CNN including its weights but does not recognize that we are using this defense method. The other is that our results indicate that the attacks function mostly by perturbing non-salient pixels, which will be a clue to understand how the CNNs yield an accurate depth map from a single (clean) image.

{\small
\bibliographystyle{ieee_fullname}
\bibliography{egbib}
}

\end{document}